\documentclass[10pt,twocolumn,letterpaper]{article}

\usepackage{wacv}
\usepackage{times}
\usepackage{epsfig}
\usepackage{graphicx}
\usepackage{amsmath}
\usepackage{amssymb}
\usepackage{adjustbox}

\usepackage{subfigure}

\usepackage{multirow}
\usepackage{textcomp}
\usepackage{colortbl}
\usepackage{xcolor}
\usepackage{array,tabularx}
\usepackage{enumitem}
\usepackage{adjustbox}

\def\wacvPaperID{428} 

\wacvfinalcopy 

\ifwacvfinal
\def\assignedStartPage{9876} 
\fi

\ifwacvfinal
\usepackage[breaklinks=true,bookmarks=false]{hyperref}
\else
\usepackage[pagebackref=true,breaklinks=true,colorlinks,bookmarks=false]{hyperref}
\fi

\ifwacvfinal
\else
\fi

\begin{document}


\title{A Study on Trees's Knots Prediction from their Bark Outer-Shape}

\author{Mohamed Mejri$^{1}$,Antoine Richard $^{2}$\\
School of Electrical and Computer Engineering, Georgia Institute of Technology, Atlanta, USA$^{1,2}$\\
{\tt\small Mohamed.mejri@gatech.edu}\\
{\tt\small Antoine.richard@gatech.edu}
\\
\and
C\'edric Pradalier$^{3}$\\
UMI2958 GT-CNRS, Metz, France$^{3}$\\
{\tt\small cedric.pradalier@georgiatech-metz.fr}
}

\maketitle

\begin{abstract}
 In the industry, the value of wood-logs strongly depends on their internal structure and more specifically on the knots' distribution inside the trees. As of today, CT-scanners are the prevalent tool to acquire accurate images of the trees internal structure. However, CT-scanners are expensive, and slow, 
 making their use impractical for most industrial applications. Knowing where the knots are within a tree could improve the efficiency of the overall tree industry by reducing waste and improving the quality of wood-logs by-products. In this paper we evaluate different deep-learning based architectures to predict the internal knots distribution of a tree from its outer-shape, something that has never been done before. Three types of techniques based on Convolutional Neural Networks (CNN) will be studied. 
 The architectures are tested on both real and synthetic CT-scanned trees. With these experiments, we demonstrate that CNNs can be used to predict internal knots distribution based on the external surface of the trees. The goal being to show that these inexpensive and fast methods could be used to replace the CT-scanners.
 Additionally, we look into the performance of several off-the-shelf object-detectors to detect knots inside CT-scanned images. This method is used to autonomously label part of our real CT-scanned trees alleviating the need to manually segment the whole of the images.

\end{abstract}

\section{Introduction}
Knots are one of the most important factors in the wood-processing chain-value. A knot is the portion of a branch embedded in the trunk that usually originates at the trunk's pith~\cite{article2}. The knots' size, shape and distribution depend on the tree specie and are influenced by environmental factors~\cite{article}.
For both forestry experts and saw-millers, knowing the exact knot distribution inside wood-logs is crucial as it directly impacts the value of the logs. When processing trees, the most valuable parts are long straight continuous portion of wood without knots because these chunks are used to make wood planks. The rest of the trees are usually processed to make small wood pieces used in furniture or shavings that are later used to make either paper or chipboard wood. Their values are significantly inferior to wood planks.

Nowadays, CT-scanners are the way to go to acquire accurate representation of the trees internal density. Yet, these machines are prohibitively expensive and take a long time to image a whole tree. Furthermore, extracting the knots out of the CT-scanned images is still an active field of research: modern methods struggle to extract knots out of all tree species.
This motivates our work to predict the knot distribution inside a tree from its outer-shape which can be acquired using fast and cheap sensors such as 2D-laser scanners. A representation of the tree's outer surface is fed to the network that outputs a representation of the knots. The choice for data representation depends on the network's architecture. Such learning is possible thanks to the strong correlation between surface defects on the tree's bark and the tree's internal knot distribution as observed in~\cite{MORISSET2012127}.
Based on this observation, we first extract the knots from the CT-scanned images using an object detection algorithm. Once we have extracted the knots from the CT-scanned images, we train various neural network architectures to predict the knots distribution using the outer-shape of trees. More precisely, three different data representation will be studied. First, we study the 2D architectures: they take as an input a patch of the log's surface map and output the density of the wood-log on the plane normal to the surface. Then we use 3D architectures: they take as an input a voxel that represents the whole log's surface and output a voxel that contains the knots. Finally, we try Convolutional-LSTM (CLSTM) networks, the input is similar to the 3D networks but the data is read layer by layer starting from the top of the log to its bottom.
To the best of our knowledge, it is the very first time that a model is used to predict the internal knot distribution from the outer-shape of a tree. 
In contrast, current methods are applied solely on CT-scanned images to detect the knots. 
 
The challenge of predicting the knots from the outer-shape arises from the high variability of the surface defects that indicates the presence of knots underneath them. Hence, to predict the knots distribution of a wood-log from its external shape, deep learning algorithms require a large dataset with many real CT-scanned logs and their corresponding outer-shape, and labeled knots. However, this data is expensive and difficult to acquire. Thus, we propose to pre-train our models on a synthetic dataset and then fine-tune them on a small dataset acquired from real-trees. 
To acquire the real dataset we use a hybrid method. We first label some knots out of the real CT-scanned images, then we train an object detector on it and use the said object detector to detect knots in the remaining CT-scanned images. Furthermore, we compare how different state-of-the-art detectors fare on this sample limited task.

In the end, this paper's contributions are the following:
\begin{itemize}
    \item Using Convolutional Neural Networks (CNN), we demonstrate that we can predict the internal knot density of a tree from its external shape. Something that has never been done before.
    \item We evaluate different data representations to find the ones that are best suited to solve the aforementioned task.
    \item We evaluate various object detectors in a sample limited scenario to detect knots inside CT-scanned images. 
\end{itemize}
\section{Related work}
\subsection{Detecting Knots from CT-Scanned Images}
Detecting knots inside CT-scanned cross-sections of logs is an active field of research, where the availability of the data and the high variability in the knots appearance are key challenges. ~\cite{JOHANSSON2013238} and~\cite{Longo2019ValidationOA} propose a method to automatically detect knots inside the heartwood of Scandinavian Scots pine (Pinus sylvestrisL.), Norwayspruce (Picea abies(L.) Karst.) and Douglas-fir (Pseudotsuga menziesii (Mirb.) Franco) logs. In \cite{JOHANSSON2013238} they suggest to model the knots by non-concentric ellipses inside the log:
First, they try to fit all the knots (i.e, a region with low-density value) inside CT-scanned Roundwood to ellipses. Then they tracked those ellipses across the cross-section of the tree until reaching the tree bark.\cite{Adrien, krahenbuhl:hal-01265531} exploited the geometrical properties of the wood combining discrete connected component extractions.
However, these methods heavily rely on user expertise to adjust the parameters of the algorithms correctly. Additionally, they are only valid on two tree species, the Scandinavian Scots Pine and the Douglas-fir.
~\cite{KnotPlanks} showed that deep learning techniques and specifically convolutional neural networks outperformed a commercial detector based on feature descriptors and SVMs 
when it comes to detecting knots in oak trees planks.
Following the observation made by~\cite{KnotPlanks}, we decided to use neural networks architectures instead of relying on the traditional computer vision methods usually used in this field to locate the knots inside the CT-scanned wood-logs.
\subsection{Object detection algorithms}
As of today, a large variety of deep-learning based object detectors can be used to find knots inside CT-scanned cross sections.
Among those, we find two-stages object detectors such as Fast R-CNN~\cite{Girshick15} and Faster-RCNN~\cite{8825470}; who first propose a set of regions of interest in which deep feature maps are cropped and classified. Although those algorithms achieve good performance, they are time-consuming due to their cascaded pipeline. Unlike two-stage detectors, a single-stage detector is faster and hence more adapted to real-time object detection purposes. The Single Shot Detector (SSD) algorithm introduced by~\cite{LiuAESR15} is one of them. It consists of a pre-trained backbone network to extract a compressed feature map and an SSD head for predicting the offsets to default boxes of different scales and aspect ratios and their associated confidences. For all these approaches, a wide range of neural network architecture can be used as a backbone structure. These include Resnet50~\cite{HeZRS15} and mobilenet~\cite{HowardZCKWWAA17}, among others. 
Another well known single stage object-detector is yolo family ~\cite{RedmonDGF15,RedmonF16,abs-1804-02767,bochkovskiy2020yolov4}. This family of models offers a good trade-off between accuracy and inference time thanks to its relatively simple decoder. Something that could be interesting in our case as often simpler architectures are less prone to over-fitting and exhibit better generalization capacities.
Finally, the current state of the art network: RetinaNet-500~\cite{abs-1708-02002} uses a FCN~\cite{LongSD14} followed by a ResNet to generate rich, multi-scale feature maps.\\
Extracting the knots from CT-scanned wood-logs is important: it helps us build our dataset, yet, predicting the knot distribution of the log from its external shape stands as our main goal. 
\subsection{Pixel-wise regression methods}
Predicting the knots distibution based on three outer-shape can be seen as a pixel wise regression task. 
In this paper, we will focus on 2-D, 3-D, and convolutional long short term memory (CLSTM) based encoder-decoders. However, in our case, the encoder will find defects on the outer-shape of the tree, and the decoder will reconstruct the inside of the tree.

\subsubsection{2D encoder-decoder structures}
2D Encoder-Decoder architectures such as SegNet~\cite{badrinarayanan2015segnet}, and U-Net~\cite{ronneberger2015unet} take an image as an input, and, when used for pixel-wise regression, output a matrix of floats of same height and width as the input. These architectures are comprised of a contractive path, (the encoder) and an expansive path (the decoder).

Fully Convolutional Networks such as DeepLab~\cite{chen2018encoderdecoder}, FCN~\cite{long2014fully}, and  PSPNet~\cite{zhao2016pyramid} rely on a very similar concept, however, the decoding process is slightly different. The decoding is done using a spatial pooling pyramid, instead of the expansive path. This results in higher accuracy in pixel-wise classification but not in pixel-wise regression as we will show later. Indeed, due to the upsampling mechanism used in the Spatial Pyramidal Pooling, the regression results are not as good as SegNet~\cite{badrinarayanan2015segnet}'s or U-Net~\cite{ronneberger2015unet}'s results.
\subsubsection{3D encoder-decoder structures}
With the advent of 3D data acquisition techniques, especially in medical imaging, several 3D neural network architectures have been designed to perform volumetric segmentation/regression. 3D-U-Net~\cite{iek20163d} is designed to perform kidney~\cite{iek20163d} segmentation. It extends the U-Net architecture~\cite{ronneberger2015unet} by replacing all 2D operations with their 3D counterparts.
~\cite{chen2016voxresnet} introduced a voxel-wise residual neural network based on residual neural network~\cite{he2015deep}. It consists of three stacked residual modules followed by four 3D-deconvolutional layers. According to~\cite{chen2016voxresnet}, VoxResNet achieves better results  than 3D-U-Net~\cite{iek20163d} after being tested on MICCAI MRBrainS challenge data~\cite{chen2016voxresnet}.

While those technics have their own advantages they each suffer of a major drawback. The 2D encoder-decoder architectures cannot capture the correlation between the successive cross sections of a log. And the volumetric segmentation/regression networks are heavily bottlenecked by their huge memory requirements which imposes low resolution input and output voxels.

\subsubsection{Convolutional-LSTM based encoder-decoder structures}
An alternative that alleviates both of these issues is to leverage the advantage of the 2D encoders: their high resolution, and add a mean for them to process sequences of input. A natural solution emmerges: Recurrent Neural Network and, more specifically, Long Short Term Memory (LSTM) structures. This family of network is widely used in natural language processing to achieve sequence-to-sequence processing.
Those neural networks aim to retrieve correlations between words in sentences.~\cite{nabavi2018future} introduces Convolutional-LSTMs (CLSTM), and  
~\cite{shi2015convolutional}) based encoder-decoder structure.\\ The encoder structure generates feature-maps from the input images that are fed to the CLSTM module. It consists of a Long Short Term Memory module where the weights are replaced with a filter bank of a convolutional layer. The decoder is composed of several deconvolutional~\cite{dumoulin2016guide} layers and combines the outputs of different CLSTM modules and generates the segmentation map for the next time-step. 
Bidirectional CLSTM based segmentation structures have also been introduced and aim to capture the temporal information in both directions.

In the following sections, we are going to test and evaluate all the aforementioned architectures to predict the internal knot distribution of the log from its outer-shape.

\section{Method}
Properly predicting the knots distribution inside trees is a challenging task due to the limited amount of CT-scanned wood-log publicly available, and the complexity of the logs' internal structures. Hence, correctly extracting the knots from the CT-scanned images is essential.

\subsection{Feature identification and extraction}
\subsubsection{Internal Knot labeling from CT-scanned cross sections}
According to forestry experts, a wide range of knots exists with different spatial behaviours and shape~\cite{10.1007/978-3-642-33191-6_21}. In this paper, only the knots that start from the pith center and extends to bark will be considered because of their direct relationship with the tree bark's shape. To assess which detector is best suited to label CT-scanned wood-logs, various object detection algorithms will be trained on a labeled subset(30$\%$) of real CT-scanned wood-logs of four different tree species: fir, elm, ash, and aspen trees. The knots have been modeled as ellipses within the labeled bounding boxes. A Faster R-CNN~\cite{8825470} with an Inception-ResNet-v2 and a RetinaNet-500~\cite{abs-1708-02002} have been selected as the best detectors based on their performance on the MS-COCO detection challenge. They were trained on a previously annotated CT-scanned subset(30$\%$ of the total dataset) of wood-logs of different tree species. This model is then used to perform coarse automatic-labeling on the 70$\%$ remaining CT-scanned cross-section images. Although Object detectors can localize the knots inside the wood-log by identifying their main features, they do not consider the correlation of the position of the knots between successive cross-section images. Object detectors can localize accurately the knots inside wood-log. However, round-woods usually contain more than one knot, and one of them is often overshadowed by other defects due to its relatively small size or low density. From a biological point of view, knots  are a continuous defect across the headwood and hence their position might be tracked along the round-woods. Therefore, we proposed to use tracking methods(e.g,CSR algorithms) to detected missed knots. Furthermore, to guarantee consistency in knots size along the log's height, a median filter is applied, so that very small or huge ellipses are filtered.
\subsubsection{External surface extraction from CT-scanned cross sections}
To extract the tree's external shape, we used a modified version of Canny algorithm\cite{4767851}.
Common edge detection algorithms such as Canny combined with filtering and other processing are often used to detect edges in Computer Vision. However, applying that algorithm failed to extract only the tree bark shape since it considers both the tree bark and the growth rings of the log as a potential edge. A solution to this is a thresholding along different axes crossing the center of the tree to extracts the shape of the tree bark accurately.\\
The extraction of the surfaces is challenging due to the high noise level that is formed by cracking and splitting of the tree bark, small branches,etc. Hence, we decided to use an object detection neural network to extract relevant extrusion on the bark surface and that might be an extension of a knot inside the log. Unlike the knots extraction that requires the tracking process, the position of the external defects remains relatively unchanged when moving throughout the cross-sections.  

\subsection{wood-log density prediction from tree bark shape}

As our current database of real CT-scanned logs is limited,
we proposed to design a look-like CT-scanned dataset based on mathematical knot modelling~\cite{andreu2003modeling}. All the deep learning based model used to predict the internal density of the logs based on the their external shape will be first be applied on our synthetic dataset to assess the performance of the models and to pretrain them on a large scale dataset. To predict the internal density we propose 3 different strategies that we present in the following subsubsection.

\subsubsection{2D neural network architectures}
      \begin{figure}[thpb]
        \centering
        \includegraphics[width = \linewidth]{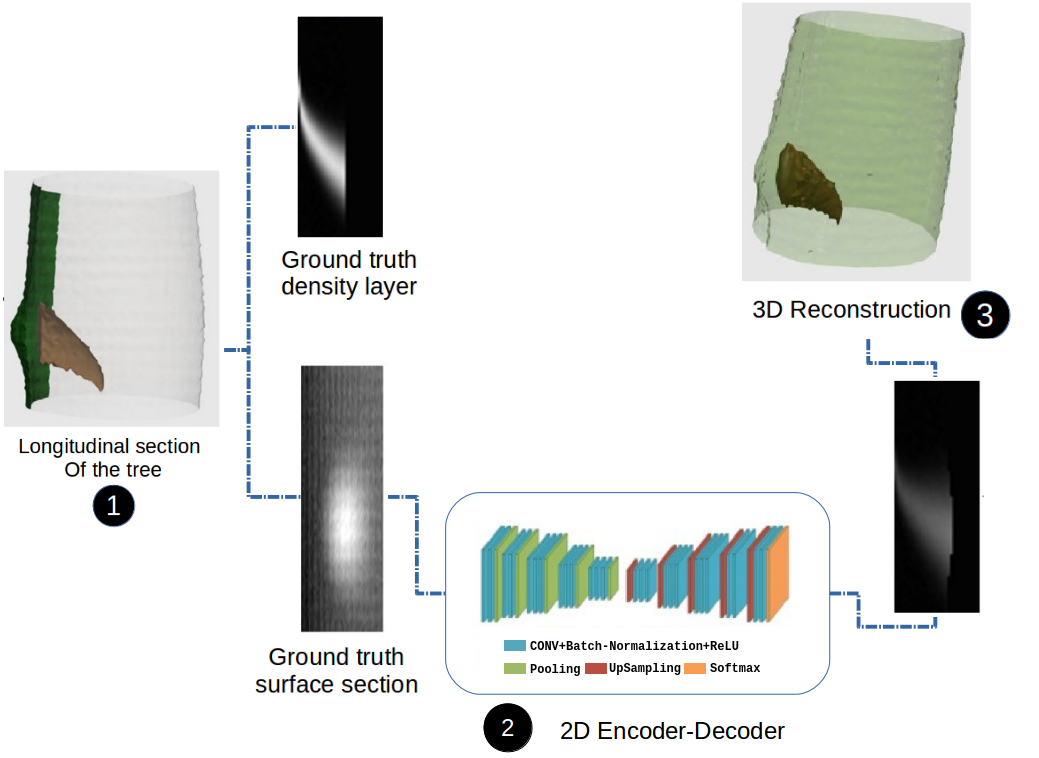}
        \caption{\small Different steps for 2D prediction and 3D reconstruction}
        \label{fig:2d2d}
    \end{figure}
    
The first strategy relies on predicting a slice of the density $d = f(r,\theta=\theta_{0},z)$ within the radial longitudinal plane using its correspondent surface $s=S(\theta \in [\theta_{0}-\frac{\pi}{12},\theta_{0}+\frac{\pi}{12}],z)$ using auto-encoders like architectures. The figure ~\ref{fig:2d2d} illustrates the prediction pipeline used in image-to-image predictions.    
As stated in the previous section, three variety of 2D-Encoder decoder will be used: SegNet~\cite{badrinarayanan2015segnet}, U-Net~\cite{ronneberger2015unet}, and DeepLabV3+~\cite{chen2018encoderdecoder}.
To use SegNet architecture as a pixel-wise regression encoder-decoder, we use an $L2$ loss(mean-squared error). Furthermore, we change the depth of the encoder and the decoder of the original SegNet due to the low-resolution input to prevent the model from overfitting. We remove the last and the first convolutional block respectively from the encoder and the decoder architectures. Additionally, we replace the rectified linear function (ReLU) with parametric-rectified linear function $P-ReLU$\cite{he2015delving} which is a learned-slope value version of leaky-ReLU. It improves model convergence while avoiding over-fitting issues. SegNet is known to be a very Deep-Network, often in Deep Networks, part of the information is lost due to a large amount of operation the data has to go through. This can results in degraded performance. 
Hence, we teste other encoder-decoder architectures that feature skip-connections which alleviate the very-deep networks issue.
U-Net\cite{ronneberger2015unet} is one of them. Here, due to low-resolution constraints, the original contracting and expanding path of the U-Net\cite{ronneberger2015unet} architecture are reduced (i.e., the 1024 filters convolution layer is removed).
The DeepLabV3+~\cite{chen2018encoderdecoder} is a revisited version of DeepLab~\cite{chen2016deeplab} architecture containing the following structures: Atrous Spatial Pooling Pyramids, Xception blocks,Depth-wise separable convolutions. 
In the original implementation of DeepLabV3+~\cite{chen2018encoderdecoder}, the L2 regularization strategy aims to avoid overfitting with a weight decay of $4e-5$. Due to the instability issue while using DeepLabV3+~\cite{chen2018encoderdecoder} for regression, we add dropout at the end of all the convolution layers instead of performing L2 regularization. This modification help stabilize the network's training.

\subsubsection{3D neural network architectures}
In this section, we present various 3D architectures, they take as an input the outer-shape of the tree as a voxel, and they output the knot distribution as a voxel. The three dimensions of the voxels represent the Cartesian coordinates x, y, and z. For the input the value of a cell is the distance between that cell and the center of the tree. For the output the cells values are the internal density. 

For this data reprentation we test three different deep learning models: 3D-U-Net~\cite{iek20163d}, 3D-SegNet and VoxResNet~\cite{chen2016voxresnet}. 
3D-SegNet is inspired by 2D-SegNet~\cite{badrinarayanan2015segnet}: the 2D layers have been replaced with their 3D counterpart and  the encoder/decoder have been shorten to reduce the memory, time expenses, and prevent overfitting. More precisely, in comparison to the original SegNet who is based on VGG-16~\cite{simonyan2014deep} the last three convolution layers whose depth are 512 have been removed in both the encoder and the decoder.
As in the original SegNet architecture, a Batch-normalization layer, and a Parametric-ReLU activation follow each convolution layer. 
To prevent overfitting, a dropout of 0.1 is added at the end of each convolution layer.

Regarding 3D-U-Net~\cite{iek20163d} we kept the original structure but made some changes in the loss function and overfitting strategy. The original 3D-U-Net~\cite{iek20163d} is trained on partially annotated voxels. Hence, the loss function has to be weighted with zero weight when the data is not annotated and one otherwise. In our case, we are performing regression with a fully-annotated dataset; we use unweighted mean-squared as a loss function. Furthermore, to prevent overfitting, a dropout of 0.1 is added after each 3D-convolution layer. 

One advantage of VoxResNet~\cite{chen2016voxresnet} over 3D-U-Net~\cite{iek20163d}  is the ability to build deeper encoder and decoder networks while avoiding overfitting. We do not make a significant change in the original architecture. We remove the final n-classes projection layer and keep the multi-level contextual information consisting of 4 3D-deconvolution layers with respectively 1,2,4 and 8 degrees of strides. Dropout of 0.1 is added after each convolutional layer to prevent overfitting.

\subsection{CLSTM based network architectures}
The CLSTMs take as an input a sequence of cross section containing the outer-shape of the tree and output a cross section with the predicted knot inside. We have modified the original architecture due to computational limitation. The SegNet~\cite{badrinarayanan2015segnet} structure is used to build the encoder and the decoder of this model. To capture the correlation between the different cross-sections of the internal density of the log while reducing the computation costs, we only keep the CLSTM\cite{shi2015convolutional} layer at the end of the encoder (bottleneck). To avoid overfitting, a dropout of 0.1 is added after each convolutional layer and inside the CLSTM Layer. 

Intuitively the correlation between a successive cross-section of log exists in forward and backward directions. To capture the correlation in both directions, the bidirectional CLSTM version of SegNet~\cite{badrinarayanan2015segnet} could be useful. It has almost the same architecture as the original CLSTM based SegNet. The main difference consists of a Bidirectional layer that replaces the original CLSTM layer. Finally, to prevent overfitting, we decided to increase the dropout from 0.1 to 0.2 inside the CLSTM layer.

\section{Experiments}
\subsection{Dataset}
\subsubsection{Synthetic look-like CT-scanned dataset}
The dataset is composed of 1800 synthetic logs. For the 3D network and the CLSTMs the inputs have a size of 256x256x64. For the 2D networks the input size is 64x64. While it may seems that the resolution of the 2D network data is lower it is not: the input of the 2D network is an angular slice of 15 degrees of the trees' surface.

\begin{figure}[thpb]
        \centering
        \includegraphics[width = 0.8\linewidth,height=2cm]{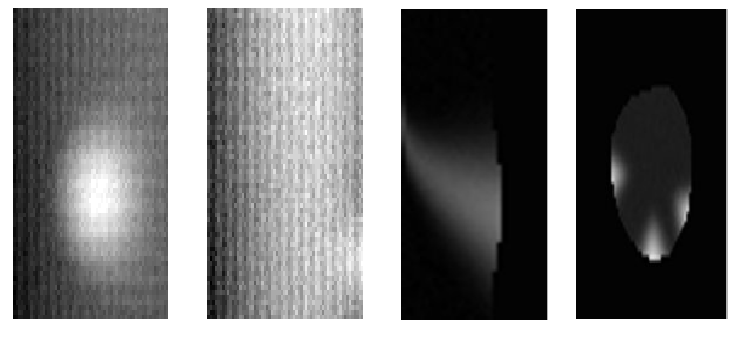}
        \caption{\small From left to right: surface with a branch, surface without branch, longitudinal section of a log, cross section of a log }
        \label{fig:syntheticimages}
    \end{figure}

The dataset is balanced in terms of the number of knots per logs. It is composed of 300 k-knots logs where k varies between 2 and 7.        
Based on the observations made in \cite{article8}, we chose to represent the tree surface as a function $s(r,\theta,z)$ where $\theta$ and $r$ refer to the polar coordinate of the tree and $z$ is the height.
To add texture onto our synthetic tree we add high and low-frequencies cosines. This creates a pattern somewhat similar to an oak bark. The longitudinal shape of the log is modeled as a decreasing function starting from its basis. Furthermore, The branches, from which the knots originates are modeled as a 2D-Gaussian function.
We model the internal density of the log as a function $d(r,\theta,z)$. The branch, a region with high density, is modeled as a combination of square root functions and linear functions. The figure ~\ref{fig:syntheticimages} illustrates two different sections of a log, as well as two external surfaces projection.   
      \\
We provide codes and scripts for reproducibility.\footnote{~\url{https://github.com/WACV2021SyntheticTree/synthetic-dataset.git}}

\subsubsection{Real CT-scanned wood-log dataset}
The dataset consist of four log of different species: fir,elm,ash and aspen tree. Each log is comprised of a sequence of 831 to 862 512x512 high resolution CT-scanned cross-section images. Each CT-scanned image is translated and cropped so that the pith center and the image center coincide. Then each wood-log image is smoothed through a 3x3 kernel size median filter to remove the noise generated by the acquisition process. 
\subsection{Network training and validation}
\begin{table*}[h]
\centering
\caption{Mean Average Prediction results of the different object detection algorithms on four tree species. Higher is better.}
\label{tab:mAP1}
\begin{center}
\begin{tabular}{c|c|c|c|c|c|c|c|c|}
    \cline{2-9} & \multicolumn{2}{c|}{Ash tree} & \multicolumn{2}{c|}{Fir tree} & \multicolumn{2}{c|}{Elm tree} & \multicolumn{2}{c|}{Aspen tree} \\ \hline
    \multicolumn{1}{|c|}{Object detector} & $mAP_{50}$         & $mAP_{75}$         & $mAP_{50}$         & $mAP_{75}$         & $mAP_{50}$         & $mAP_{75}$         & $mAP_{50}$          & $mAP_{75}$          \\ \hline
    \multicolumn{1}{|c|}{Faster R-CNN~\cite{8825470}}    & 35.2          & 34.7          & 34.2          & 32.7          & 36.0          & 33.4          & 30.3           & 28.7           \\ \hline
    \multicolumn{1}{|c|}{SSD-513~\cite{LiuAESR15}}         & 25.9          & 24.7          & 26.9          & 23.7          & 23.6          & 22.7          & 21.4           & 20.1           \\ \hline
    \multicolumn{1}{|c|}{RetinaNet-500~\cite{abs-1708-02002}}   & \bf43.7          & 40.9          & \bf42.7          & 41.9          & \bf44.3          & 41.2          & \bf36.9           & 34.7           \\ \hline
    \multicolumn{1}{|c|}{YoloV3-608~\cite{abs-1804-02767}}      & 37.4          & 32.6          & 36.4          & 33.6          & 38.9          & 36.3          & 31.0           & 29.8           \\ \hline
\end{tabular}
\end{center}
\end{table*}
\subsubsection{Object Detection}
To extract the knots and the external defect from the CT-scanned images
we chose to use object detection algorithms provided by the Tensorflow object detection API~\cite{HuangRSZKFFWSG016}. It allows to easily reproduce our experiments and provides a diverse set of contemporary 
architectures~\footnote{~\url{https://github.com/tensorflow/models/tree/master/research/object_detection}}. Each detector (i.e. SSD~\cite{LiuAESR15}, Faster RCNN and RetinaNet-500~\cite{abs-1708-02002}) are pre-trained on MS COCO dataset and fine-tuned on 32000 annotated synthetic CT-scanned images for 60 epochs.
\subsubsection{Internal Knots Prediction}
To predict the knot distribution of the log from its outer-shape, we split the synthetic dataset into three subsets: training set, validation set, and testing set with the respective proportion $80\%$, $4\%$, and $16\%$.\\
The hyperparameter of the 2D and 3D predictors are tuned as follows: we chose an Adam optimizer with an initial learning rate of $10^{-3}$, a learning rate scheduler is also used to avoid undesirable divergent behavior while increasing the number of epochs(the learning rate drops to $10^{-4}$ after 20 epochs). 
For computational constraints, the batch size is fixed to two voxels, 100 images, and a sequence of 64 images for respectively the 3D models, 2D models, and CLSTM based models. Each model is trained for 50 epochs.  At the end of each epoch, a the models are evaluated on the validation set and we save only the weights of the model with a minimum validation error. 
\subsubsection{Evaluation Metrics}

To evaluate the performances of the object detecion algorithms we use the most common metrics: the mean Average Precision (mAP) with two thresholds: $50\%$ and $75\%$.\\
Quantitative and qualitative metrics were used to assess the quality of the internal density prediction. The Root means square error (RMSE) of the difference voxel to voxel between the ground truth and the predicted volumetric structure is applied to evaluate the performance of both the 3D and the CLSTM based encoder-decoder models. It was also used to assess the prediction quality of the 2D models after 3D reconstruction of the ground truth and the predicted logs.

\section{Results}
\subsection{Internal and external knots extraction}
In this section, we assess and compare the results of the different object detectors applied on our four real CT-scanned trees. 
Table ~\ref{tab:mAP1} shows the mAP of the different detectors used to detect the knots as well as the branches(i.e., external defects) with and without the tracking process.

Table ~\ref{tab:mAP1} shows the average precision of the four different object detection algorithms on high resolution CT-scanned wood-log (i.e. 512x512). Compared to Faster detectors like SSD-513~\cite{LiuAESR15} or YoloV3-608~\cite{abs-1804-02767}, RetinaNet-500~\cite{abs-1708-02002} consitently achieves more accurate results in detecting the knots of all the tree species. Even though, RetinaNet-500~\cite{abs-1708-02002} is more time consuming and computationally expensive its superior performances make it the most relevant choice to extract the knots and branches.
\begin{table*}[h]
\caption{RMSE of the different type of architectures : On the synthetic dataset. Lower is better.} \label{tab:SRMSE}
\begin{center}
\begin{tabular}{|l||c|c|c|c|}\hline
    \multirow{2}{*}{Architecture} & \multicolumn{3}{c|}{RMSE $(10^{-2})$} & \multirow{2}{*}{Parameters} \\
    \cline{2-4} & 2 $Branches$ & 5 $Branches$ & 7 $Branches$ & \\\hline\hline
    SegNet~\cite{ronneberger2015unet}                        & \bf1.27       & \bf1.66       & 2.46       & 34 M                        \\ \hline
    U-Net                         & 1.33       & 1.68       & \bf2.42       & 36 M                        \\ \hline
    DeepLabV3+~\cite{chen2018encoderdecoder}                    & 3.17       & 3.23       & 3.48       & 42 M                        \\ \hline\hline
    3D-SegNet & 3.42 & 3.68 & 4.22 & 144 M \\\hline
    3D-U-Net~\cite{iek20163d} & 3.53 & 3.72 & 4.30 & 114 M \\\hline
    VoxResNet~\cite{chen2016voxresnet} & 3.6 & 3.92 & 4.2 & 35 M\\\hline\hline
    CLSTM-SegNet & 3.61 & 3.89 & 4.26 & 31 M \\\hline
    Bidir-CLSTM-SegNet & \textbf{3.1} & \textbf{3.29} & \textbf{3.9} & 52 M \\\hline
\end{tabular}
\end{center}
\end{table*}
When looking at the results, the difference betzeen the mAP50 and mAP75 is minimal meaning that the objects that the object detected by the detectors are well detected but it misses quite a lot of them. While it's far from being an autonomous solution it can be used as a prelabeling tool where the user then adds or remove the incorrect detection. Moreover, all of the object detection algorithms have a relatively high false negatives rate when it comes to distinguish between knots that extends to a branch and small surface defects generated by a cracking inside the log. Hence, we decided to keep only trackable defects that extends to the tree bark and could be predicted from the outer-shape of the log. 
As explained earlier, our current database of real CT-scanned trees is extremely limited. As of today  only four trees are available. This explains why the detectors are struggling, yet RetinaNet clearly shows that it is at least the most sample efficient architecture if not simply the most accurate architecture. This is extremely important as the database that we are currently building will not exceed hundred of trees.

\begin{figure}[thpb]
\begin{center}
    \hfill
\subfigure[\centering \tiny(Upper Left) Elm , (Upper Right) ash, (Lower Left) aspen  (Lower Right) fir ]{\includegraphics[width =0.45\linewidth,height=4cm]{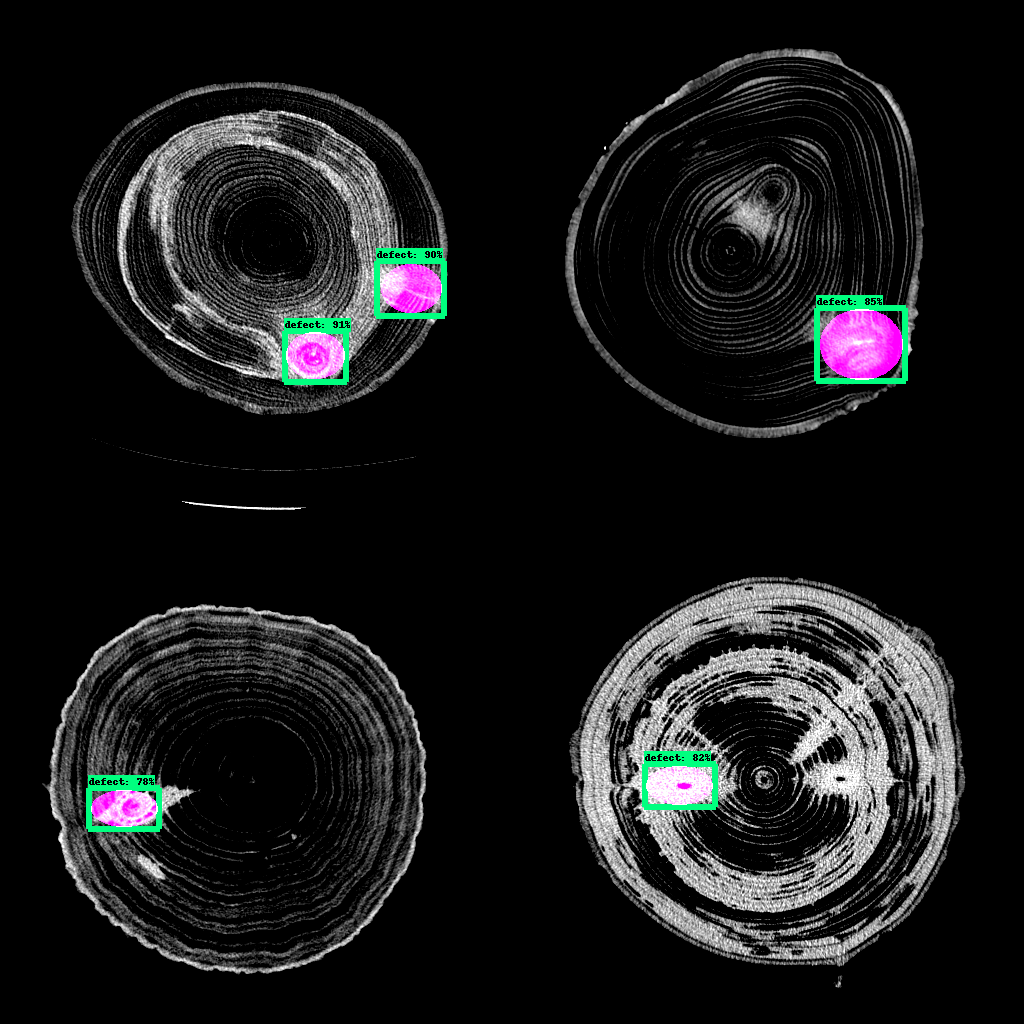}}
\hfill
\subfigure[\centering \tiny (Upper Left) Elm , (Upper Right) ash , (Lower Left) fir  (Lower Right) aspen ]{\includegraphics[width = 0.45\linewidth,height=4cm]{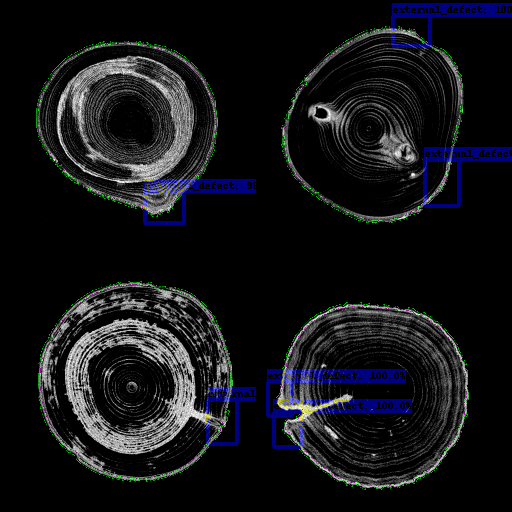}}
\hfill
\caption{(Left) Retina-Net detected bounding boxes and their corresponding ellipses representing knots.(Right)  Retina-Net detected bounding boxes of external branches and epicormic shoots }
\end{center}
\label{fig:elmfirashaspen}
\end{figure}

\subsection{Knots Distribution Prediction}
\subsubsection{Synthetic Dataset}
In this section, we will assess the knots prediction based on 288 synthetic log external surfaces. Table ~\ref{tab:SRMSE} shows the mean performance of the different 2D, 3D, and CLSTM based neural network architectures when tested on 40 k-knots logs ($k\in \{2,5,7\}$). Each density prediction model has been trained five times, and we compute the average RMSE to get accurate results.

From the results shown in table ~\ref{tab:SRMSE}, it can be seen that the 2D networks largely outperform their 3D counterparts. Within the 2D networks, SegNet achieves better prediction results than DeepLabV3+~\cite{chen2018encoderdecoder} and U-Net for two branches logs. In contrast, U-Net achieves better results for predicting the internal density of logs with a high number of branches. Those results are expected because, unlike the SegNet model, U-Net architecture focuses more on capturing the information lost through the encoder structure, which may be relevant when it comes to more complex log structure (i.e., logs with a high number of branches). DeepLabV3+ fails compared to the other architectures to achieve good results: The DeepLabV3+ architecture features four times upsampling at the end of the network that may explain why its performances are lower than the other networks. \\When looking at the voxel-based networks, 3D-SegNet and 3D-U-Net~\cite{iek20163d} achieve better performance than VoxResNet~\cite{chen2016voxresnet}. Unlike 3D-U-Net and VoxResNet, 3D-SegNet doesn't contain skip-connections; hence it may be affected by information loss within the encoding process, as seems to suggest the decrease in performance of the network as the number of branches increases.
Finally, the sequence-based models offered the best performance of all the 3D models on the synthetic dataset; this remains true as the number of branches increases. 
Moreover, the LSTMs have fewer parameters than the other 3D models making them particularly interesting with real data as they will have less chances of overfitting.


All in all, on the synthetic data, the 2D approaches performed better than their 3D counterparts. This can be explained by the different types of inputs the 2D networks and the voxel-based/sequence-based networks have. Indeed to predict the internal density, the 2D networks rely on a slice of the tree surface, which in our configuration yields an angular resolution of about 4.2 pixels per degree. On the other hand, the 256x256 voxels offer, at most, the equivalent of an angular resolution of 2.2 pixels per degree. This drastic difference in resolution explains why the 2D networks performed better. As of today the voxel-based networks would have to be able to handle much larger voxels to match the resolution of the 2D nets, which is not possible yet on our 16Gb GPUs.

Unfortunately, the 2D networks cannot be applied to the real trees, as they need a high-resolution surface map that we cannot generate from the CT-scanned trees. The CT-scanned images at our disposal do not have the resolution and finesse required to create proper surface maps of the trees. 
However, they should be applicable to datasets that features both CT-scans and lidar acquired external tree shapes. A dataset that is currently under construction.

\begin{table*}[h]
\caption{RMSE of the 3D architectures applied on real trees. Lower is better. }
\label{tab:RMSE}
\begin{center}
\begin{tabular}{|l||c|c|c|c|}

\hline 
\multirow{2}{*}{Architecture} & \multicolumn{4}{c|}{RMSE $(10^{-2})$} \\ \cline{2-5}
                              & Fir tree & elm tree& ash tree & aspen tree                             \\ \hline\hline
3D-SegNet                     & 5.98       & 6.13       & 7.92       & 6.51                        \\ \hline
Bidir-CLSTM-SegNet            & 4.56       & 4.4        & 5.23       & 4.1                        \\ \hline

\end{tabular}
\end{center}
\end{table*}
\subsubsection{Real Dataset}
The prediction of the knots distribution in the real wood-logs is more challenging due to the few samples available and the high variability of the surface patterns indicating the presence of knots beneath them.
This observations might explain the results shown in table ~\ref{tab:RMSE}.
Like in the previous experiment, the CLSTMs show superior performances when compared to 3D SegNet, a voxel-based network. While these results are not as good as on the synthetic dataset, it shows that the CLSTMs are not only more accurate but also remains more accurate with a limited dataset. 

The CLSTMs results are encouraging as they were acquired using a single real-log in the fine-tuning process. As shown in \ref{fig:voxelreal}, the results on the fir and elm trees are particularly promising.

\subsection{3D visual results}
\begin{figure}[thp]
        \centering
        \includegraphics[width = \linewidth,height=3cm]{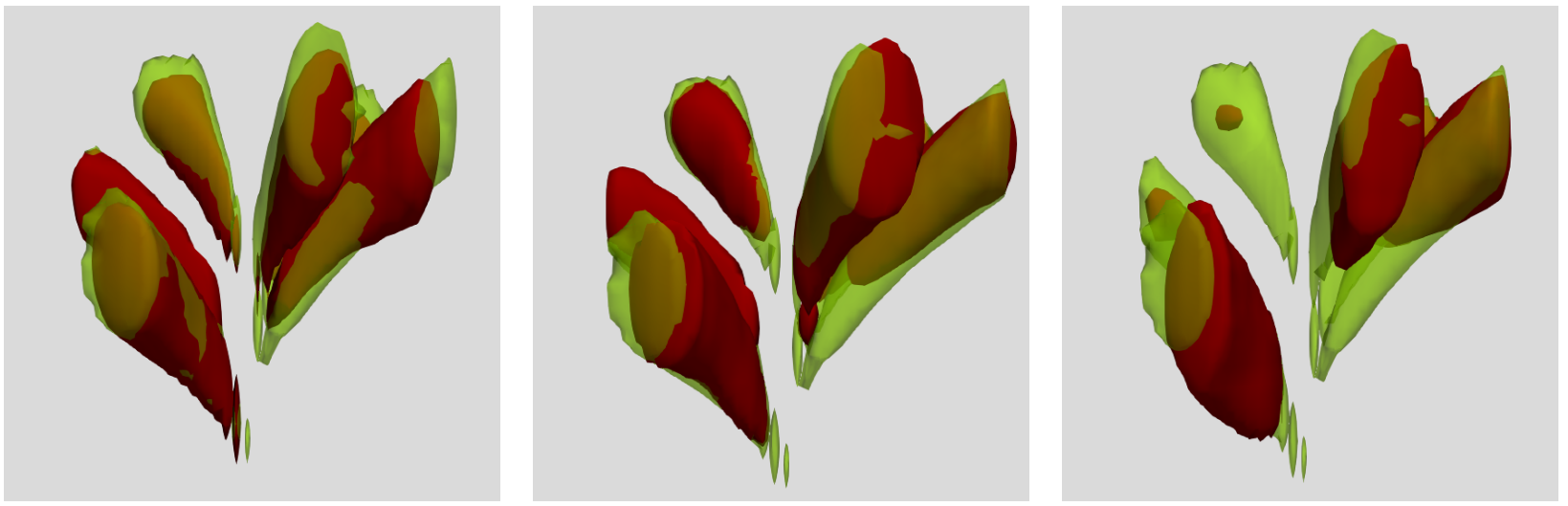}
        \caption{\small From left to right: 3D-U-Net~\cite{iek20163d}, U-Net, Bidirectional CLSTM based SegNet}
        \label{fig:syntheticvoxel}
    \end{figure}

We used  $Paraview^{\copyright}$, an open-source multi-platform application for 3D visualization, to assess the 3D rendering of the predicted iso-surface of the internal density of the tree. Figure ~\ref{fig:syntheticvoxel} shows the predicted iso-surface with different models for the 6-branches synthetic log. Based on the RMSE evaluation, we chose to assess only the 2D, 3D, and CLSTM based models with the highest performance. We conclude that the 2-D models achieve better results than 3-D models in terms of iso-surface correspondence between the ground truth (Green volume) and the predicted log (Red volume).    
    
\begin{figure}[thp]
        \centering
        \includegraphics[width = \linewidth,height=3.5cm]{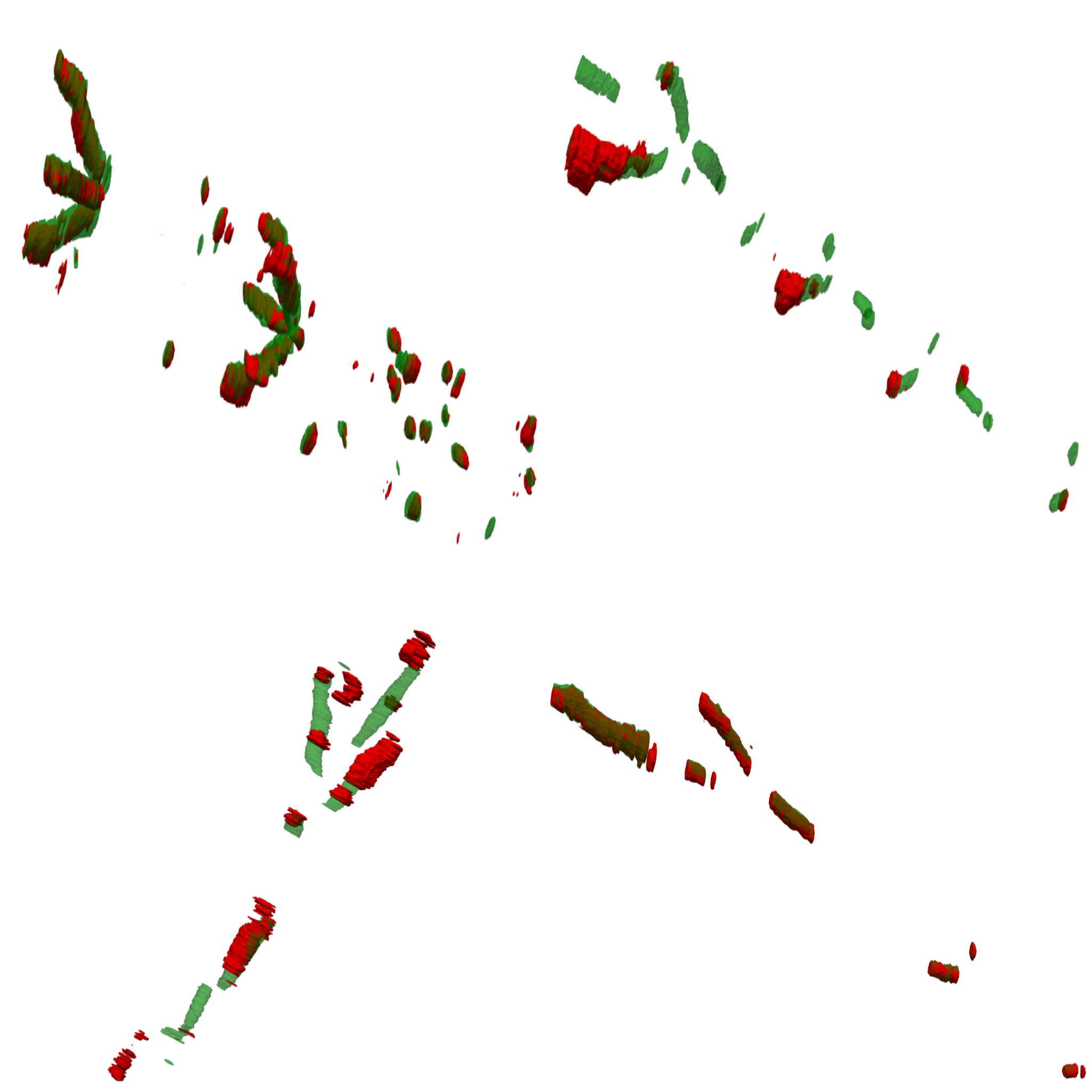}
        \caption{\small (Upper Left) fir tree, (Upper Right) ash tree, (Lower Left) aspen tree (Lower Right) elm tree}
        \label{fig:voxelreal}
    \end{figure}

As stated in the previous section, only predicted iso-surface derived from real CT-scanned wood-log using pre-trained Bidir-CLSTM~\cite{shi2015convolutional} SegNet and 3D-Seg-Net will be visually assessed. The figure ~\ref{fig:voxelreal} shows a volumetric representation of the predicted internal density of the trees using a CLSTM and their corresponding ground truth.  
It is easier for the neural network architectures to predict knots closer to the bark tree (ex: elm tree) than those closer to the pith (ex: ash tree).

\section{Conclusion}
In this paper, and for the first time, we show that the internal knots distribution can be inferred by CNNs from the outer-shape of the trees. In doing so we study the performances of a wide diversity of neural networks. First to automatically label the CT-scanned cross-section images using a object detectors. To do so, we leveraged manually annotated images and extended our current dataset by applying object-detectors on the remaining non-labeled images. Amongst the evaluated object detectors RetinaNet performed best even though the dataset was limited. Then we evaluated 3 different data representations on both synthetic and real data. Amidst the different data representation applied on the synthetic dataset the 2D one performed best. The 3D CNNs and the CLSTM falling far behind. However, the 2D representation could not be tested on the data we currently have at our disposal, hence the real data was tested only on the CLSTMs and 3D CNNs. On the real dataset like on the synthetic one, the best performing model were the bi-directionnal CLSTMs. Overall, the performances of the CLSTMs model on the real trees, are not that different from the synthetic trees indicating that those results should be applicable to the larger and more sophisticated dataset that we are currently acquiring. This first proof of concept comfort us our belief that CT-scanners are not the only solution to infer the internal structure of trees. In future work we will apply CLTMs and 2D CNNs to a larger and more diverse dataset. And finally, we will improve the realism of our synthetic dataset by leveraging GANs. 

{\small
\bibliographystyle{ieee_fullname}
\newpage
\bibliography{egbib}

\begin{thebibliography}{10}\itemsep=-1pt

\bibitem{andreu2003modeling}
Jean-Philippe Andreu and Alfred Rinnhofer.
\newblock Modeling knot geometry in norway spruce from industrial ct images.
\newblock In {\em Scandinavian Conference on Image Analysis}, pages 786--791.
  Springer, 2003.

\bibitem{article8}
Jean-Philippe Andreu and Alfred Rinnhofer.
\newblock Modeling of internal defects in logs for value optimization based on
  industrial ct scanning.
\newblock pages 23--26, 2003.

\bibitem{badrinarayanan2015segnet}
Vijay Badrinarayanan, Alex Kendall, and Roberto Cipolla.
\newblock Segnet: A deep convolutional encoder-decoder architecture for image
  segmentation.
\newblock 2015.

\bibitem{bochkovskiy2020yolov4}
Alexey Bochkovskiy, Chien-Yao Wang, and Hong-Yuan~Mark Liao.
\newblock Yolov4: Optimal speed and accuracy of object detection.
\newblock 2020.

\bibitem{4767851}
J. {Canny}.
\newblock A computational approach to edge detection.
\newblock {\em IEEE Transactions on Pattern Analysis and Machine Intelligence},
  PAMI-8(6):679--698, 1986.

\bibitem{chen2016voxresnet}
Hao Chen, Qi Dou, Lequan Yu, and Pheng-Ann Heng.
\newblock Voxresnet: Deep voxelwise residual networks for volumetric brain
  segmentation.
\newblock 2016.

\bibitem{chen2016deeplab}
Liang-Chieh Chen, George Papandreou, Iasonas Kokkinos, Kevin Murphy, and
  Alan~L. Yuille.
\newblock Deeplab: Semantic image segmentation with deep convolutional nets,
  atrous convolution, and fully connected crfs.
\newblock 2016.

\bibitem{chen2018encoderdecoder}
Liang-Chieh Chen, Yukun Zhu, George Papandreou, Florian Schroff, and Hartwig
  Adam.
\newblock Encoder-decoder with atrous separable convolution for semantic image
  segmentation.
\newblock 2018.

\bibitem{article}
Emmanuel Duchateau, David Auty, Frederic Mothe, Longuetaud Fleur, C. Ung, and
  Alexis Achim.
\newblock Models of knot and stem development in black spruce trees indicate a
  shift in allocation priority to branches when growth is limited.
\newblock {\em PeerJ}, 3, 04 2015.

\bibitem{dumoulin2016guide}
Vincent Dumoulin and Francesco Visin.
\newblock A guide to convolution arithmetic for deep learning.
\newblock 2016.

\bibitem{Girshick15}
Ross~B. Girshick.
\newblock Fast {R-CNN}.
\newblock {\em CoRR}, abs/1504.08083, 2015.

\bibitem{HeZRS15}
Kaiming He, Xiangyu Zhang, Shaoqing Ren, and Jian Sun.
\newblock Deep residual learning for image recognition.
\newblock {\em CoRR}, abs/1512.03385, 2015.

\bibitem{he2015deep}
Kaiming He, Xiangyu Zhang, Shaoqing Ren, and Jian Sun.
\newblock Deep residual learning for image recognition.
\newblock 2015.

\bibitem{he2015delving}
Kaiming He, Xiangyu Zhang, Shaoqing Ren, and Jian Sun.
\newblock Delving deep into rectifiers: Surpassing human-level performance on
  imagenet classification.
\newblock 2015.

\bibitem{HowardZCKWWAA17}
Andrew~G. Howard, Menglong Zhu, Bo Chen, Dmitry Kalenichenko, Weijun Wang,
  Tobias Weyand, Marco Andreetto, and Hartwig Adam.
\newblock Mobilenets: Efficient convolutional neural networks for mobile vision
  applications.
\newblock {\em CoRR}, abs/1704.04861, 2017.

\bibitem{HuangRSZKFFWSG016}
Jonathan Huang, Vivek Rathod, Chen Sun, Menglong Zhu, Anoop Korattikara,
  Alireza Fathi, Ian Fischer, Zbigniew Wojna, Yang Song, Sergio Guadarrama, and
  Kevin Murphy.
\newblock Speed/accuracy trade-offs for modern convolutional object detectors.
\newblock {\em CoRR}, abs/1611.10012, 2016.

\bibitem{8825470}
L. {Jiao}, F. {Zhang}, F. {Liu}, S. {Yang}, L. {Li}, Z. {Feng}, and R. {Qu}.
\newblock A survey of deep learning-based object detection.
\newblock {\em IEEE Access}, 7:128837--128868, 2019.

\bibitem{JOHANSSON2013238}
Erik Johansson, Dennis Johansson, Johan Skog, and Magnus Fredriksson.
\newblock Automated knot detection for high speed computed tomography on pinus
  sylvestris l. and picea abies (l.) karst. using ellipse fitting in concentric
  surfaces.
\newblock {\em Computers and Electronics in Agriculture}, 96:238 -- 245, 2013.

\bibitem{krahenbuhl:hal-01265531}
Adrien Kr{\"a}henb{\"u}hl, Bertrand Kerautret, and Isabelle Debled-Rennesson.
\newblock {TKDetection: a software to detect and segment wood knots}.
\newblock {\em {imagen-a}}, 3(5), Mar. 2013.

\bibitem{10.1007/978-3-642-33191-6_21}
A. Kr{\"a}henb{\"u}hl, B. Kerautret, I. Debled-Rennesson, F. Longuetaud, and F.
  Mothe.
\newblock Knot detection in x-ray ct images of wood.
\newblock In George Bebis, Richard Boyle, Bahram Parvin, Darko Koracin,
  Charless Fowlkes, Sen Wang, Min-Hyung Choi, Stephan Mantler, J{\"u}rgen
  Schulze, Daniel Acevedo, Klaus Mueller, and Michael Papka, editors, {\em
  Advances in Visual Computing}, pages 209--218, Berlin, Heidelberg, 2012.
  Springer Berlin Heidelberg.

\bibitem{Adrien}
Adrien Krähenbühl, Bertrand Kerautret, and Isabelle Debled-Rennesson.
\newblock Knot segmentation in noisy 3d images of wood.
\newblock volume 7749, pages 383--394, 03 2013.

\bibitem{abs-1708-02002}
Tsung{-}Yi Lin, Priya Goyal, Ross~B. Girshick, Kaiming He, and Piotr
  Doll{\'{a}}r.
\newblock Focal loss for dense object detection.
\newblock {\em CoRR}, abs/1708.02002, 2017.

\bibitem{LiuAESR15}
Wei Liu, Dragomir Anguelov, Dumitru Erhan, Christian Szegedy, Scott~E. Reed,
  Cheng{-}Yang Fu, and Alexander~C. Berg.
\newblock {SSD:} single shot multibox detector.
\newblock {\em CoRR}, abs/1512.02325, 2015.

\bibitem{LongSD14}
Jonathan Long, Evan Shelhamer, and Trevor Darrell.
\newblock Fully convolutional networks for semantic segmentation.
\newblock {\em CoRR}, abs/1411.4038, 2014.

\bibitem{long2014fully}
Jonathan Long, Evan Shelhamer, and Trevor Darrell.
\newblock Fully convolutional networks for semantic segmentation.
\newblock 2014.

\bibitem{article2}
Bruna Longo, Franka Brüchert, Gero Becker, and Udo Sauter.
\newblock Validation of a ct knot detection algorithm on fresh douglas-fir
  (pseudotsuga menziesii (mirb.) franco) logs.
\newblock {\em Annals of Forest Science}, 76, 06 2019.

\bibitem{Longo2019ValidationOA}
Bruna~L. Longo, Franka Br{\"u}chert, Gero Becker, and Udo~H. Sauter.
\newblock Validation of a ct knot detection algorithm on fresh douglas-fir
  (pseudotsuga menziesii (mirb.) franco) logs.
\newblock {\em Annals of Forest Science}, 76:1--16, 2019.

\bibitem{MORISSET2012127}
Jean-Baptiste Morisset, Frédéric Mothe, and Francis Colin.
\newblock Observation of quercus petraea epicormics with x-ray ct reveals
  strong pith-to-bark correlations: Silvicultural and ecological implications.
\newblock {\em Forest Ecology and Management}, 278:127 -- 137, 2012.

\bibitem{KnotPlanks}
Rickard Norlander, Josef Grahn, and Atsuto Maki.
\newblock Wooden knot detection using convnet transfer learning.
\newblock pages 263--274, 06 2015.

\bibitem{RedmonDGF15}
Joseph Redmon, Santosh Divvala, Ross Girshick, and Ali Farhadi.
\newblock You only look once: Unified, real-time object detection.
\newblock 2015.

\bibitem{RedmonF16}
Joseph Redmon and Ali Farhadi.
\newblock {YOLO9000:} better, faster, stronger.
\newblock {\em CoRR}, abs/1612.08242, 2016.

\bibitem{abs-1804-02767}
Joseph Redmon and Ali Farhadi.
\newblock Yolov3: An incremental improvement.
\newblock {\em CoRR}, abs/1804.02767, 2018.

\bibitem{ronneberger2015unet}
Olaf Ronneberger, Philipp Fischer, and Thomas Brox.
\newblock U-net: Convolutional networks for biomedical image segmentation.
\newblock 2015.

\bibitem{nabavi2018future}
Seyed shahabeddin Nabavi, Mrigank Rochan, Yang, and Wang.
\newblock Future semantic segmentation with convolutional lstm.
\newblock 2018.

\bibitem{shi2015convolutional}
Xingjian Shi, Zhourong Chen, Hao Wang, Dit-Yan Yeung, Wai kin Wong, and Wang
  chun Woo.
\newblock Convolutional lstm network: A machine learning approach for
  precipitation nowcasting.
\newblock 2015.

\bibitem{simonyan2014deep}
Karen Simonyan and Andrew Zisserman.
\newblock Very deep convolutional networks for large-scale image recognition.
\newblock 2014.

\bibitem{zhao2016pyramid}
Hengshuang Zhao, Jianping Shi, Xiaojuan Qi, Xiaogang Wang, and Jiaya Jia.
\newblock Pyramid scene parsing network.
\newblock 2016.

\bibitem{iek20163d}
Özgün Çiçek, Ahmed Abdulkadir, Soeren~S. Lienkamp, Thomas Brox, and Olaf
  Ronneberger.
\newblock 3d u-net: Learning dense volumetric segmentation from sparse
  annotation.
\newblock 2016.

\end{thebibliography}
}
\end{document}